\documentclass[12pt, a4paper]{article}

\usepackage[utf8]{inputenc}
\usepackage[T1]{fontenc}
\usepackage{microtype}              % For better typography
\usepackage[margin=1in]{geometry}   % Set page margins
\usepackage{amsmath}                % For advanced math environments
\usepackage{amssymb}                % For math symbols
\usepackage{graphicx}               % To include images
\usepackage{booktabs}               % For professional-quality tables
\usepackage{tabularx}               % For tables with auto-wrapping text
\usepackage{authblk}                % For author and affiliation blocks
\usepackage{algorithm}              % For algorithm floats
\usepackage{algpseudocode}          % For pseudocode layout
\usepackage{xcolor}                 % For custom colors
\usepackage{wrapfig}
\usepackage{hyperref}               % For hyperlinks (crucial for modern documents)

\usepackage{natbib} % 加载natbib以获得灵活的引用命令
\definecolor{darkblue}{rgb}{0.0, 0.0, 0.55}
\definecolor{darkred}{rgb}{0.55, 0.0, 0.0}

\hypersetup{
    colorlinks=true,
    linkcolor=darkblue,
    citecolor=darkred,
    filecolor=magenta,
    urlcolor=cyan,
    pdftitle={Socratic RL: A Novel Framework for Efficient Knowledge Acquisition},
    pdfauthor={Xiangfan Wu},
}

\title{
    \hrule
    \vspace{0.4em}
    \textbf{Socratic RL: A Novel Framework for Efficient Knowledge Acquisition through Iterative Reflection and Viewpoint Distillation}
    \vspace{0.4em}
    \hrule
}
\author{
    Xiangfan Wu \\
    \affil{Ocean University of China}
}
\date{\today}

\begin{document}

\maketitle

\begin{abstract}
Current Reinforcement Learning (RL) methodologies for Large Language Models (LLMs) often rely on simplistic, outcome-based reward signals (e.g., final answer correctness), which limits the depth of learning from each interaction. This paper introduces Socratic Reinforcement Learning (Socratic-RL), a novel, process-oriented framework designed to address this limitation. Socratic-RL operates on the principle that deeper understanding is achieved by reflecting on the causal reasons for errors and successes within the reasoning process itself. The framework employs a decoupled "Teacher-Student" architecture, where a "Teacher AI" analyzes interaction histories, extracts causal insights, and formulates them into structured "viewpoints." These viewpoints, acting as distilled guidance, are then used by a "Student AI" to enhance its subsequent reasoning. A key innovation is the iterative self-improvement of the Teacher AI, enabling its reflective capabilities to evolve through a meta-learning loop. To manage the accumulation of knowledge, a distillation mechanism compresses learned viewpoints into the Student's parameters. By focusing on process rather than just outcome, Socratic-RL presents a pathway toward enhanced sample efficiency, superior interpretability, and a more scalable architecture for self-improving AI systems. This paper details the foundational concepts, formal mechanisms, synergies, challenges, and a concrete research roadmap for this proposed framework.
\end{abstract}

\section{Introduction}
Reinforcement learning (RL) has emerged as a key technology for fine-tuning large language models (LLMs) ~\cite{shao2024deepseekmath, guo2025deepseek, liu2025understanding, achiam2023gpt, team2025kimi, zhang2025right, lin2025cppo, xiong2025minimalist, zhang2025right, yuan2025vapo, hu2025open,xu2025not} to improve their reasoning and accuracy on complex tasks. Leading systems, such as OpenAI's GPT-4o and o1~\cite{openai2024o1}, Google's Gemini~\cite{team2023gemini}, Anthropic's Claude 3 Opus~\cite{anthropic2024claude}, and DeepSeek~\cite{deepseek2024v2,shao2024deepseekmath,guo2025deepseek}, all leverage RL techniques to enhance their capabilities beyond what is possible with supervised learning alone. In domains requiring sophisticated reasoning, RL serves as the core mechanism driving their remarkable performance.

However, prevailing RL methodologies for LLMs often rely on simplistic and outcome-oriented reward signals (e.g., final answer correctness). This coarse-grained feedback mechanism significantly limits the depth and breadth of learning from each interaction.

To address this limitation, this paper introduces Socratic Reinforcement Learning (Socratic-RL), a novel, process-oriented framework. Its core principle is that deeper understanding is achieved by reflecting on the causal chain of successes and failures within the reasoning process itself, rather than merely observing the final outcome. Socratic-RL employs a decoupled ``Teacher-Student'' architecture. A ``Teacher AI'' is tasked with analyzing interaction histories, extracting the underlying causal relationships for errors and successes, and formulating them into structured ``viewpoints.'' These viewpoints, serving as distilled guidance, are then used by a ``Student AI'' to guide and enhance its subsequent reasoning.

A key innovation of the framework is the iterative self-improvement of the Teacher AI. Through a meta-learning loop, the Teacher's reflective and analytical capabilities are designed to evolve and strengthen over time. To manage the continuous accumulation of knowledge, a knowledge distillation mechanism is introduced to efficiently compress and integrate the learned viewpoints into the Student AI's parameters.

By focusing on the process rather than merely the outcome, Socratic-RL presents a pathway toward AI systems with enhanced sample efficiency, superior interpretability, and a more scalable architecture for self-improvement. This paper provides a detailed account of the framework's foundational conc

\section{Related Work and Theoretical Foundations}
\paragraph{Outcome vs. Process Supervision.} A cornerstone of our work is the distinction between Outcome Supervision and Process Supervision. While Reinforcement Learning from Human Feedback (RLHF)~\cite{ouyang2022training,dong2024rlhf} is a form of outcome supervision, rewarding final outputs, recent work has shown the power of process supervision, which rewards the intermediate steps of a model's reasoning. Socratic-RL can be seen as a novel form of automated process supervision. Instead of requiring human annotation for each reasoning step, the Teacher AI learns to automatically generate process-level feedback (viewpoints) and improves this ability over time.

\paragraph{AI Feedback and Self-Improvement.} The framework's architecture shares a conceptual lineage with Reinforcement Learning from AI Feedback (RLAIF) ~\cite{lee2023rlaif}. However, RLAIF systems typically use a static AI critiquer. They lack a mechanism for the critiquer to learn from the outcomes of its own critiques. Socratic-RL directly addresses this gap by introducing a meta-learning loop where the Teacher AI's ability to generate insightful 'viewpoints' is itself refined based on the Student's subsequent performance, creating a system that not only learns, but learns how to teach. This also differentiates it from frameworks like Self-Refine~\cite{madaan2023self}, where the same model performs both generation and critique. By decoupling the Teacher and Student, Socratic-RL allows for specialized optimization: the Student becomes an expert at solving tasks, while the Teacher becomes an expert at reflection and causal analysis.

\paragraph{Efficiency in RL Fine-Tuning.}  Our work is also situated within a growing body of research focused on improving the efficiency of RL fine-tuning. Methods like GPRO~\cite{shao2024deepseekmath} and its variants aim to optimize the learning process at a granular level by identifying and selectively updating only the most salient sub-networks or parameters (e.g., LoRA modules) for a given task. This focus on \textit{parameter-level saliency}, which identifies \textit{where} in the network to apply updates, is complementary to Socratic-RL's approach. Instead of optimizing the mechanics of the gradient update, Socratic-RL focuses on \textit{semantic-level saliency}. It identifies the most critical conceptual flaw in a reasoning trace and generates a high-level "viewpoint" as feedback. In essence, while GPRO makes the learning process more efficient by optimizing the \textit{update mechanism}, Socratic-RL aims to achieve efficiency by improving the \textit{quality and abstraction of the learning signal itself}, addressing the \textit{why} of the error, not just the \textit{how} of the correction.

\paragraph{Prompting, Distillation, and Meta-Learning.} The "viewpoints" generated by our Teacher AI are a form of dynamic, contextualized Prompting. Unlike static, manually-crafted prompts, viewpoints are automatically generated based on a deep analysis of past performance, targeting specific identified weaknesses in the Student's reasoning. Furthermore, the knowledge compression cycle is a direct application of Knowledge Distillation (KD), and the evolving Teacher AI firmly places our framework within the domain of Meta-Learning ("learning to learn"). By integrating these diverse concepts, Socratic-RL aims to create a more holistic, efficient, and intelligent learning paradigm.

\section{The Socratic-RL Framework: Architecture and Formalism}

\begin{figure}[h!]
    \centering
    \includegraphics[width=\textwidth]{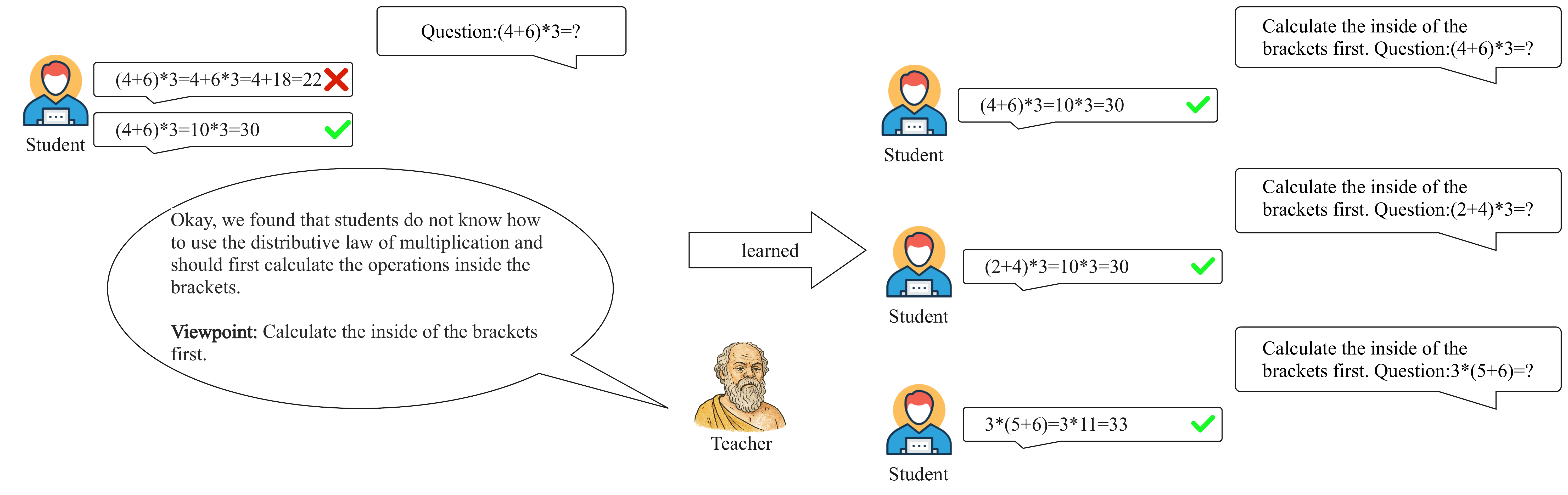}
    \caption{A high-level overview of the Socratic-RL framework.}
    \label{fig:socratic_rl}
\end{figure}

\subsection{System Overview and Formalism}

The Socratic-RL framework is conceptualized as a bi-level optimization process involving two primary agents: a Student AI tasked with solving problems and a Teacher AI tasked with generating pedagogical feedback. The system's objective is to iteratively refine both agents to maximize the Student's ultimate task performance. We formalize the core components in the context of autoregressive language models, where tasks involve generating a sequence of tokens $y = (y_1, \dots, y_T)$.

\begin{itemize}
    \item \textbf{State and Action Space:} In the LLM context, a state $s_t$ represents the partial context for generation at step $t$, comprising the initial prompt $x_{\text{prompt}}$ and the sequence of previously generated tokens: $s_t = (x_{\text{prompt}}, y_1, \dots, y_{t-1})$. The action $a_t$ corresponds to generating the next token, $y_t$. The full generated sequence is the trajectory of actions.

    \item \textbf{Environment Interaction Trace ($\tau$):} A trace is a complete record of the Student's interaction with a task. It is a sequence $\tau = (s_0, a_0, \dots, s_T, a_T, R)$, where $s_0$ is the initial prompt, $a_T$ is the final action (e.g., an end-of-sequence token), and $R$ is a scalar reward signal received upon completion. For complex reasoning tasks, intermediate rewards are typically zero ($r_0, \dots, r_{T-1} = 0$), and only a final outcome-based reward $R$ is provided, making credit assignment a significant challenge.

    \item \textbf{Student Policy ($\pi_S$):} The Student AI is an autoregressive policy, parameterized by $\theta_S$, which generates an action (token) given a state and a set of active viewpoints. Its objective is to maximize the expected final reward. The policy is formally written as $\pi_S(a_t | s_t, V; \theta_S)$, where $V$ is the set of active viewpoints, typically prepended to the input context $s_t$.

    \item \textbf{Viewpoint ($v$):} A viewpoint is a piece of structured, human-readable text representing a generalizable principle, a heuristic, a causal explanation, or a counter-example. It is designed to be a portable piece of knowledge that can guide the Student's reasoning process. For instance, a viewpoint could be: \textit{"Principle: In multi-step arithmetic, always resolve expressions within parentheses before applying external operators."}

    \item \textbf{Teacher Policy ($\pi_T$):} The Teacher AI is a generative model, parameterized by $\theta_T$, that takes a full interaction trace $\tau$ as input and performs a form of automated causal analysis. It aims to identify the root cause of failure (or success) within the Student's reasoning trace and synthesizes this insight into a new viewpoint $v$. The policy is thus defined as $v \sim \pi_T(\tau; \theta_T)$. The Teacher's goal is not to solve the task itself, but to produce viewpoints that are maximally effective for improving the Student's policy $\pi_S$.
\end{itemize}

The overarching goal is to find optimal parameters $\theta_S^*$ for the Student. This is achieved by iteratively improving the Teacher's parameters $\theta_T$ such that the viewpoints it generates accelerate the learning of the Student. The system thus solves a meta-learning problem where the Teacher learns how to teach effectively.

\subsection{The Core Loop: From Interaction to Viewpoint}
The Student AI interacts with the environment. The Teacher AI observes the Student's interaction trace $\tau$ to perform a causal analysis, identifying specific failure modes like \textbf{logical fallacies, procedural errors, or flawed assumptions}. Based on this analysis, the Teacher generates a "viewpoint." For example, if a Student fails `(4 + 6) * 3` by calculating `4 + 18`, the Teacher might generate the viewpoint: \textit{"In arithmetic, operations inside parentheses must be evaluated first."} This viewpoint is then added to the set of active viewpoints $V$ to guide the Student.

\begin{algorithm}[H]
\caption{The Socratic-RL Core Loop}
\label{alg:socratic-rl}
\begin{algorithmic}[1]
\State Initialize Student policy $\pi_S$ and Teacher policy $\pi_T$.
\State Initialize an empty set of active viewpoints $V = \emptyset$.
\State Initialize a knowledge base of all learned viewpoints $\mathcal{V}_{KB} = \emptyset$.
\For{each episode $k = 1, 2, ...$}
    \State \Comment{\textit{Phase 1: Student Interaction}}
    \State Sample a task and generate an interaction trace $\tau_k \sim \pi_S(\cdot | s, V)$.
    \State Obtain outcome (e.g., success/failure, final reward $R_k$).
    \State
    \State \Comment{\textit{Phase 2: Teacher Reflection}}
    \If{outcome is suboptimal or meets generation criteria}
        \State Generate a new viewpoint $v_k \sim \pi_T(\tau_k)$.
        \State Add $v_k$ to the active viewpoints: $V \leftarrow V \cup \{v_k\}$.
        \State Add $v_k$ to the knowledge base: $\mathcal{V}_{KB} \leftarrow \mathcal{V}_{KB} \cup \{v_k\}$.
    \EndIf
    \State
    \State \Comment{\textit{Phase 3: Meta-Learning (Teacher Evolution)}}
    \State Update $\pi_T$ using feedback on the utility of past viewpoints from $\mathcal{V}_{KB}$ (see Section 3.3).
    \State
    \State \Comment{\textit{Phase 4: Knowledge Distillation}}
    \If{a distillation condition is met (e.g., fixed interval)}
        \State Fine-tune a new Student $\pi'_S$ to internalize viewpoints in $\mathcal{V}_{KB}$.
        \State Set $\pi_S \leftarrow \pi'_S$ (the new Student becomes the current one).
        \State Reset active viewpoints: $V \leftarrow \emptyset$.
    \EndIf
\EndFor
\end{algorithmic}
\end{algorithm}

\subsection{The Meta-Learning Engine: Evolving the Teacher AI}
We posit that the Teacher AI's capabilities are context-dependent, varying across different environments. The Teacher AI's ability to generate effective "viewpoints" is a learnable skill, and its evolution is driven by a meta-objective: to learn how to generate viewpoints that construct the most effective prompts for the Student, thereby maximizing its learning progress. We formalize the quality of a viewpoint $v$ with a utility score $U(v)$, which measures the performance uplift it provides when added to the Student's context in a set of probe tasks $\mathcal{P}_{\text{probe}}$:
$$
U(v) = \mathbb{E}_{p \sim \mathcal{P}_{\text{probe}}}[\text{Score}(\pi_S(\cdot|p, V \cup \{v\}))] - \mathbb{E}_{p \sim \mathcal{P}_{\text{probe}}}[\text{Score}(\pi_S(\cdot|p, V))]
$$
Where $\text{Score}(\cdot)$ is the task success metric. Through this mechanism, we can obtain a Teacher AI that continuously improves its own prompts.

\subsection{Knowledge Distillation and Scalability}
A core challenge for scalability is that relying on an ever-growing set of explicit viewpoints $V$ in the prompt context is computationally inefficient and has a finite limit. To ensure long-term learning and create a more capable standalone model, Socratic-RL incorporates a modular knowledge distillation mechanism. This mechanism's purpose is to compress the procedural knowledge encapsulated in the viewpoint-guided interactions into the parameters $\theta_S$ of a new Student model, $\pi'_S$.

The default method for this is \textbf{policy distillation} (or behavioral cloning). Here, the new Student $\pi'_S$ is trained to mimic the output distribution of the original, viewpoint-guided Student $\pi_S$. This is achieved by minimizing the Kullback-Leibler (KL) divergence between the two policies over a dataset $\mathcal{D}$ of inputs and viewpoints from the knowledge base $\mathcal{V}_{KB}$:
$$
\mathcal{L}_{\text{distill}} = \mathbb{E}_{(\text{Input}, v) \sim \mathcal{D} \times \mathcal{V}_{KB}} \left[ D_{KL}\left( \pi_S(\cdot | \text{Input}, v; \theta_S) \parallel \pi'_S(\cdot | \text{Input}; \theta'_S) \right) \right]
$$
This effectively trains the new Student to behave as if it "knows" the principles from the viewpoints without needing to see them.

However, the distillation module is designed to be plug-and-play, allowing for more advanced techniques to be employed. Alternative strategies include:

\begin{itemize}
    \item \textbf{Direct Preference Optimization (DPO):} The viewpoints can be used to generate preference pairs. For a given problem, we can generate one response using a relevant helpful viewpoint ($y_{\text{preferred}}$) and another response without it, or with a deliberately unhelpful "negative viewpoint" ($y_{\text{rejected}}$). This creates a dataset of triplets $(x_{\text{prompt}}, y_{\text{preferred}}, y_{\text{rejected}})$ which can be used to fine-tune the Student model directly via DPO, a powerful and highly effective method.

    \item \textbf{Instruction Tuning:} The entire knowledge base of viewpoints $\mathcal{V}_{KB}$ can be systematically reformatted into a high-quality instruction-tuning dataset. Each viewpoint is transformed into a general instruction. For example, the viewpoint \textit{"Principle: In arithmetic, evaluate parentheses first."} can generate training examples like: \texttt{\{"instruction": "Solve the following, paying close attention to the order of operations.", "input": "(4+6)*3", "output": "..."\}}. This approach directly integrates the learned principles into the model's fundamental instruction-following capabilities.
\end{itemize}

The choice of distillation method is a key hyperparameter of the framework, allowing practitioners to balance between direct behavioral cloning and more nuanced preference-based or instruction-based fine-tuning to achieve optimal performance.

\subsection{Interpretability by Design}
A significant advantage of Socratic-RL is its inherent interpretability. The set of learned viewpoints, $\mathcal{V}_{KB}$, serves as a human-readable log of the system's acquired knowledge. By examining the sequence of generated viewpoints, researchers can trace the AI's learning trajectory, understand the principles it has discovered, and diagnose its failures. This "glass-box" nature contrasts sharply with the "black-box" behavior of models trained solely on outcome-based rewards. The viewpoints themselves become artifacts for analysis, representing the AI's evolving "theory" of the world.

\section{Synergies and Comparative Analysis}
The modular design of Socratic-RL allows it to be synergistically combined with other advanced AI techniques, enhancing its capabilities and safety.

\paragraph{Integration with Existing Methods.} The meta-learning loop for the Teacher AI is highly compatible with Direct Preference Optimization (DPO)~\cite{rafailov2023direct}, which can be used to refine its ability to generate helpful viewpoints. Furthermore, the Teacher's generation process can be constrained by a set of predefined rules, akin to Constitutional AI (CAI)~\cite{bai2022constitutional}, ensuring that the guidance it provides is safe, ethical, and aligned with human values. 

\paragraph{Comparative Analysis.} Table \ref{tab:comparison} provides a high-level comparison with existing LLM training paradigms, highlighting the unique combination of features in Socratic-RL.

\begin{table}[h!]
\centering
\caption{Socratic-RL vs. Existing LLM Training Paradigms}
\label{tab:comparison}
\small
\begin{tabularx}{\textwidth}{@{}lXXX@{}}
\toprule
\textbf{Paradigm} & \textbf{Primary Feedback Type} & \textbf{Source of Feedback} & \textbf{Iterative Improvement of Source} \\
\midrule
\textbf{Socratic-RL} & Structured, causal "Viewpoints" (Process) & Evolving "Teacher AI" & Yes (Core feature) \\
\addlinespace
\textbf{RLHF}~\cite{ouyang2022training} & Scalar Reward (Outcome) & Human Annotators & No \\
\addlinespace
\textbf{RLAIF}~\cite{lee2023rlaif} & Natural Language Critiques (Outcome) & AI Critiquer Model & No (Typically static) \\
\addlinespace
\textbf{DPO}~\cite{rafailov2023direct} & Preference Pairs (Outcome/Process) & Human/AI Preferences & No \\
\addlinespace
\textbf{CAI}~\cite{bai2022constitutional} & Adherence to Rules/Principles & Pre-defined Constitution & No \\
\addlinespace
\textbf{Self-Refine}~\cite{madaan2023self} & Self-Generated Critiques (Process) & Model Itself & Yes (Not specialized) \\
\bottomrule
\end{tabularx}
\end{table}

\section{Challenges and Limitations}
Despite its promise, the practical implementation of Socratic-RL faces significant hurdles that require careful consideration.

\paragraph{Subjectivity in Evaluation.} The framework's efficacy hinges on the Teacher AI's ability to discern "better" from "worse" reasoning. While straightforward in objective domains like math, this becomes ill-defined in tasks with subjective criteria, such as creative writing or summarization. Defining a suitable utility function $U(v)$ for the Teacher in such domains is a substantial open research problem.

\paragraph{Stability of Self-Improvement.} Self-referential training loops are notoriously susceptible to instability. The system could suffer from \textbf{model collapse}, where the diversity of outputs diminishes over time. Worse, it could create feedback loops that \textbf{amplify biases}. For instance, if a Teacher develops a slight stylistic preference, it may generate viewpoints that reinforce this style in the Student, which in turn provides evidence for the Teacher's preference, leading to a rapid loss of stylistic diversity.

\paragraph{Computational Cost and Latency.} The Socratic loop is computationally intensive. Each cycle involves an extra inference pass from the Teacher model, analysis of the trace, and potentially calculating the utility of viewpoints. This introduces significant latency and cost compared to standard fine-tuning, which may be a barrier to its application in resource-constrained or real-time scenarios.

\paragraph{Teacher Model Drift.} The unconstrained evolution of the Teacher AI carries the risk of "epistemic drift" or even "madness," where it might develop bizarre or unhelpful theories about the world that are locally consistent but globally incorrect. Ensuring the Teacher remains grounded in reality and aligned with the intended learning goals is a critical safety and performance challenge.

\section{Methodological Approach}
\begin{wrapfigure}{r}{0.5\textwidth}
    \vspace{-20pt} % 可选：向上移动图表，微调垂直位置
    \centering
    \includegraphics[width=\linewidth]{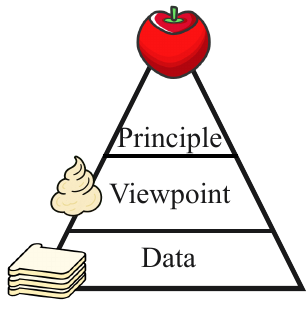} % 使用 \linewidth 可以让图片宽度自适应 wrapfigure 的宽度
    \caption{Knowledge Hierarchy in Socratic-RL: Principles, Viewpoints, and Data}
    \label{fig:data}
    \vspace{-5pt} % 可选：向下移动图表下方的文本，微调垂直位置
\end{wrapfigure}

The foundational principle of our framework is to deeply integrate core RL concepts, such as reward functions and policy gradients, directly into the optimization process of the LLM itself. This approach transforms these abstract algorithmic components into explicit, model-driven operations.

Recently, a critical aspect of RL that has gained significant attention is the concept of entropy, particularly its role in identifying the most salient parts of an experience. For instance, recent research from Alibaba has demonstrated that a minority of tokens often play a disproportionately large role during the RL process~\cite{wang2025beyond}. Similarly, frameworks like SEED-GPRO~\cite{chen2025seed} leverage the semantic entropy of a problem to determine which parameters to update, thereby focusing the learning process on areas of higher uncertainty.

Traditional methods of defining rewards and updating parameters struggle to distill abstract reasoning or "the right way of thinking" directly from input-output pairs. Our Socratic Reinforcement Learning (Socratic-RL) framework addresses this problem by empowering the LLM to take on a more direct role within the learning algorithm.

Although this adds a layer of complexity to the RL process, it leverages a unique capability of Large Language Models: the ability to increase information density. This function is key to enabling the model to rapidly acquire new knowledge and principles from a very small number of examples.

We propose a hierarchy of knowledge representation where information density progressively increases. It begins with raw text, is then refined into token dependency relationships, and is ultimately distilled into concrete "viewpoints." This progressive concentration of knowledge is the core mechanism for achieving more effective and efficient knowledge compression within the model. For example, as shown in Figure ~\ref{fig:data}, a "Principle" represents the directly inputted context. However, due to the limited input space, these are often guiding principles, such as "think step by step" or instructions in the form of an agent, which can guide general problems. Our "viewpoint" resides at an intermediate level; for a specific problem, it can provide more detailed guidance than a Principle, thereby facilitating the problem-solving process. At the same time, it is much smaller than the raw data, thus enabling the generalized resolution of similar problems.

% 'r' 表示图表在右侧 (right)。您也可以用 'l' 表示左侧 (left)。
% {0.5\textwidth} 表示侧边栏的宽度为总文本宽度的一半，您可以按需调整。
% [10pt] 是一个可选参数，表示图表外部与文本之间的额外间距。

% 结束环绕后，正文会恢复正常排版。

\section{Future Research Directions}
The challenges outlined above define a clear roadmap for future research. We propose a phased strategy to systematically validate and mature the Socratic-RL framework.

\begin{enumerate}
    \item \textbf{Phase 1: Proof-of-Concept with a Rule-Based Teacher.} The immediate next step is to implement the framework in a constrained, objective domain like multi-step arithmetic. The Teacher AI will initially be a rule-based system that parses the Student's Chain-of-Thought trace to detect specific, pre-defined errors (e.g., incorrect order of operations). Key research questions for this phase include: (a) Does viewpoint-guided learning show higher sample efficiency than outcome-based RL? (b) Is the knowledge distillation cycle effective at compressing procedural knowledge into the Student's parameters without catastrophic forgetting?

    \item \textbf{Phase 2: Developing and Training an LLM-based Teacher.} Following the proof-of-concept, we will replace the rule-based system with an LLM-based Teacher. This phase will focus on the meta-learning aspect. We will experiment with different prompt-engineering strategies for the Teacher to elicit causal analysis and will implement and compare training methods (e.g., RL with utility rewards vs. DPO on viewpoint pairs). The central goal is to demonstrate that the Teacher's ability to teach can be measurably improved through automated feedback.

    \item \textbf{Phase 3: Scaling, Stability, and Safety.} Once the core mechanics are validated, we will scale the framework to more complex domains like code generation. This phase will directly tackle the stability and safety challenges. We will investigate techniques to mitigate model collapse and bias amplification, such as using an ensemble of diverse Teachers, regularizing the Teacher's policy to prevent drastic shifts, and continuously grounding the system with a stream of fresh, external data.
\end{enumerate}

\section{Conclusion}
Socratic Reinforcement Learning offers a conceptual and architectural blueprint for a new generation of LLMs that learn more efficiently and transparently. By shifting the training paradigm from sparse, outcome-based rewards to rich, process-oriented feedback, Socratic-RL opens the door to more sample-efficient learning. The framework's core innovations—the decoupled, evolving Teacher-Student architecture and the knowledge distillation cycle—provide a promising, albeit challenging, path toward more capable, interpretable, and continuously improving artificial intelligence. Successfully addressing the outlined challenges could represent a significant step towards creating AI systems that not only solve problems, but understand the principles behind the solutions.

% --- Placeholder for Bibliography ---
% \bibliographystyle{plain}
% \begin{thebibliography}{9}
% \bibitem{example1}
% A. Author, B. Author, "Title of a related work," \textit{Journal Name}, Vol. X, pp. 1-10, 2024.
% \end{thebibliography}
\bibliographystyle{unsrt} % 指定参考文献的样式文件（例如 plainnat.bst）
\bibliography{ea} % 指定 .bib 文件的名字（不需要写 .bib 后缀）
\end{document}